\documentclass[letterpaper]{article} 
\usepackage[preprint]{aaai2027}  
\usepackage[hyphens]{url}  
\usepackage{graphicx} 
\usepackage{natbib}  
\usepackage{caption} 
\usepackage{amsmath} 
\usepackage{algorithm}
\usepackage{algorithmic}
\usepackage{multirow}
\usepackage{newfloat}
\usepackage{listings}
\DeclareCaptionStyle{ruled}{labelfont=normalfont,labelsep=colon,strut=off} 
\floatstyle{ruled}
\newfloat{listing}{tb}{lst}{}
\floatname{listing}{Listing}

\usepackage{booktabs}
\usepackage{amsmath} 
\usepackage{amsfonts} 
\usepackage{xcolor}
\usepackage{colortbl} 
\title{DeAR: Decentralized Agentic Reasoning via Capability Grounding and Collaborative Thought Navigation}
\author{
    Wei Xing\textsuperscript{\rm 1},
    Zheng Changmeng\textsuperscript{\rm 2},
    Wei Xiaoyong\textsuperscript{\rm 2},
    Ye Xiufen\textsuperscript{\rm 1},
    Li Qing\textsuperscript{\rm 2}
}

\affiliations{
    \textsuperscript{\rm 1}College of Intelligent Systems Science and Engineering, Harbin Engineering University\\
    \textsuperscript{\rm 2}Department of Computing, The Hong Kong Polytechnic University\\
    m18324618517@163.com, 
    yexiufen@hrbeu.edu.cn, 
    changmeng.zheng@polyu.edu.hk, 
    cs007.wei@polyu.edu.hk, 
    qing-prof.li@polyu.edu.hk
}

\begin{document}

\maketitle

\begin{abstract}
Existing agentic reasoning systems typically rely on centralized protocols. This design introduces routing bottlenecks and static role allocations that often fail when handling complex multimodal queries. We propose \textbf{DeAR} (\textbf{De}centralized \textbf{A}gentic \textbf{R}easoning), a framework that shifts from central control to autonomous peer-to-peer collaboration. DeAR is built on three mechanisms: (1) decentralized capability grounding for query-dependent agent specialization, (2) thought map navigation for targeted peer interactions, and (3) topology update for adaptive error correction. Evaluations across 9 diverse multimodal reasoning and text-based QA benchmarks indicate that DeAR consistently outperforms recent baseline methods, validating that decentralized and adaptive collaboration among agents enhances accuracy in knowledge-intensive reasoning tasks. The source code will be available at \url{https://open_upon_acceptance}.
\end{abstract}


\section{Introduction}
Complex reasoning represents the frontier of Large Language Model (LLM) applications, demanding capabilities that extend far beyond simple pattern matching. To address this, \textbf{Agentic Reasoning} has emerged as a proven and effective paradigm \citep{zheng2023rethinking}. By decomposing intricate problems into manageable sub-tasks handled by specialized agents, agentic workflows have achieved remarkable proficiency in multi-step problem solving \citep{zheng2024picture}.

\begin{figure}[t]
    \vspace{-0.5pt}  
    \centering  
    \includegraphics[width=1\linewidth]{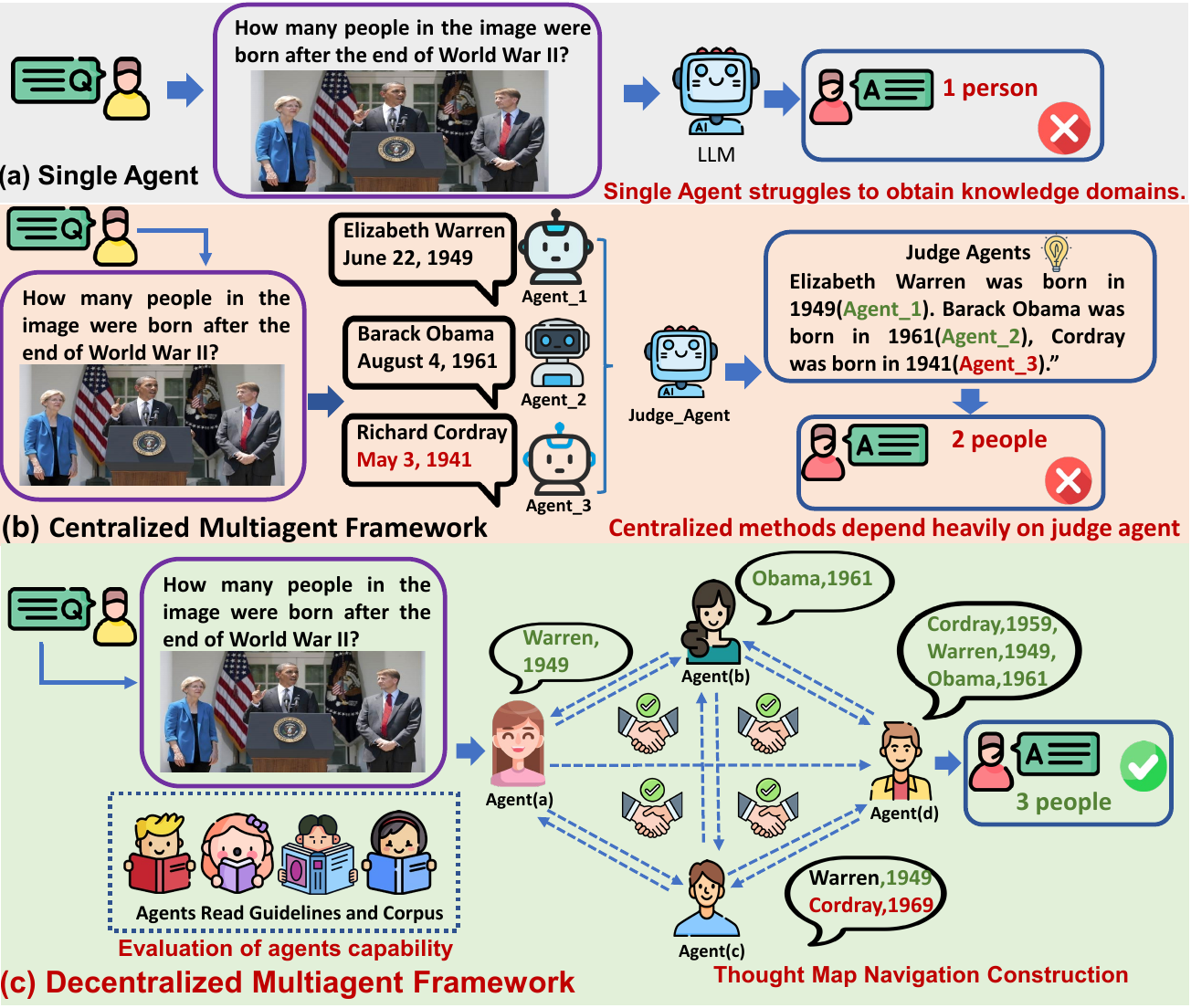}
    \caption{Framework comparison on a multimodal query ("How many people in the image were born after World War II?"). (a) Single agent fails due to restricted knowledge domains. (b) Centralized framework fails as the judge bottleneck propagates incorrect inputs. (c) DeAR (Ours) succeeds by grounding capabilities and navigating an adaptive thought map to collaboratively verify facts and yield the correct answer.}
    \label{fig:1}     
\end{figure}

Despite these successes, current agentic reasoning systems \citep{zhou2025shielda} frequently encounter significant stumbling blocks, particularly when handling complex multimodal queries. Consider the task in Figure \ref{fig:1}, which asks how many people in a specific image were born after World War II. A standard single agent often fails here because it cannot simultaneously manage fine-grained visual recognition and historical knowledge retrieval. Furthermore, in a centralized multi-agent framework, a planner rigidly assigns visual extraction to one agent and factual querying to another. If an agent misidentifies a person or hallucinates a birth year, the central "Judge" node becomes a severe bottleneck. It blindly aggregates these flawed intermediate results without peer verification, resulting in an incorrect final answer.

We argue that these failures stem from structural flaws within the Centralized Collaboration Protocol rather than deficient model capabilities. In these topologies, a central coordinator monopolizes routing and aggregation \citep{wang2024megaagent}, inducing severe information loss and strict single-point bottlenecks \citep{owens2025multi}. Consequently, this rigid role allocation \citep{jia2024agentstore} fails to adapt to fluid, query-dependent reasoning, forcing dynamic problems into static workflows that misroute tasks and propagate unverified errors downstream.

In the real world, complex problem-solving follows a different topology. It is inherently \textbf{decentralized}. Interdisciplinary experts do not wait for a single omniscient manager to dictate every micro-interaction; they engage in peer-to-peer discourse, negotiate boundaries, and self-organize to bridge knowledge gaps.

However, this decentralized paradigm remains largely unexplored in the context of agentic reasoning.To bridge this gap, we propose \textbf{DeAR} (\textbf{De}centralized \textbf{A}gentic \textbf{R}easoning), a framework designed to bypass the bottlenecks of centralization by enabling agents to reason and collaborate autonomously. The evolving reasoning context is maintained not by a hidden system layer, but as a dynamic message payload passed directly through peer-to-peer token streams. Realizing robust decentralized reasoning requires addressing three challenges, which we tackle by restructuring role allocation, collaboration mechanisms, and reasoning paths:
\begin{itemize}
    \item \textbf{Dynamic capability vs. Static role}:
    Centralized systems often rely on pre-defined roles \citep{fang2025comprehensive}, allowing the coordinator to route requests by labels.
    But role labels are a weak proxy for real competence and are precisely what causes coordinators to hallucinate capabilities in open-ended scenarios.
    We address this by replacing static labels with a \emph{dynamic capability grounding} mechanism. Agents actively assess their own suitability for a specific query fragment based on verifiable linguistic benchmarks, ensuring that role allocation is query-dependent and driven by actual competence rather than rigid titles.

    \item \textbf{Local graph navigation vs. Central routing}:
    In centralized pipelines \citep{wu2025talk}, the coordinator acts as a global router that dictates a linear execution path.
    To eliminate this central bottleneck while maintaining selective collaboration, we propose a \emph{collaboration propensity matrix}, which intuitively functions as a dynamic \textbf{adjacency matrix} defining the edges of our reasoning graph.
    Analogous to sociolinguistic alignment, this matrix establishes a decision boundary. Instead of a central judge computing a global sequential route, the currently active agent performs a \emph{local graph traversal}, independently selecting its optimal downstream peer based on the current context.

    \item \textbf{Progressive reasoning vs. Restarting}:
    In centralized systems, if a reasoning chain breaks, the ``Judge'' often has to discard the entire trajectory and restart, which is highly inefficient.
    We propose that decentralized reasoning should be viewed as collaborative navigation over an agentic thought map.
    Rather than fragile linear chains, our agents maintain a shared topological map of the reasoning space. This allows for \emph{progressive reasoning}: if a path proves invalid, agents do not retreat to the starting line. Instead, they efficiently pivot from the current node, utilizing the thought map to explore alternative effective chains without losing the progress made thus far.
    
\end{itemize}
More details are provided in Section~\ref{method}.
\section{Related work}
Recent studies show that multi-agent systems improve the robustness and interpretability of Large Language Models (LLMs) and Multimodal Large Language Models (MLLMs) by coordinating agents with complementary capabilities. These frameworks facilitate complex task decomposition, intermediate state sharing, and dynamic workflow management \citep{hong2023metagpt}. Consequently, multi-agent collaboration has been applied successfully across diverse domains, including mathematical reasoning \citep{xie2024mathlearner} and visual question answering, where iterative planning and reversible reasoning enhance overall performance \citep{xinjie2025reagent}.

To address complex multi-step reasoning tasks, recent frameworks explore distributing cognitive loads across specialized agents \citep{jiang2025towards, liu2025hm}. While debate-based approaches\citep{zheng2024picture,liang2026multi,liang2026mixture} and linear reasoning paths allow for the refinement of intermediate states \citep{hu2025removal}, they heavily rely on rigid, centralized coordination protocols. For instance, AutoGen \citep{wu2024autogen} depends on static topologies governed by central managers, AgentVerse \citep{chen2024agentverse} dictates synchronous pipelines under central evaluators, and DyLAN \citep{liu2024dynamic} requires a global ranker for layer-wise filtering. Even dynamic routing frameworks like AgentNet \citep{yang2025agentnet} lack query-dependent capability grounding and adaptive error correction. This persistent centralized bottleneck limits adaptability, incurs redundant computational overhead, and leaves systems vulnerable to hallucinated competencies and compounding reasoning errors, highlighting the critical need for a decentralized collaborative architecture.
\section{DeAR Framework}

\label{method}

\begin{figure*}[tbp] 
    \centering
    \includegraphics[width=0.9\textwidth]{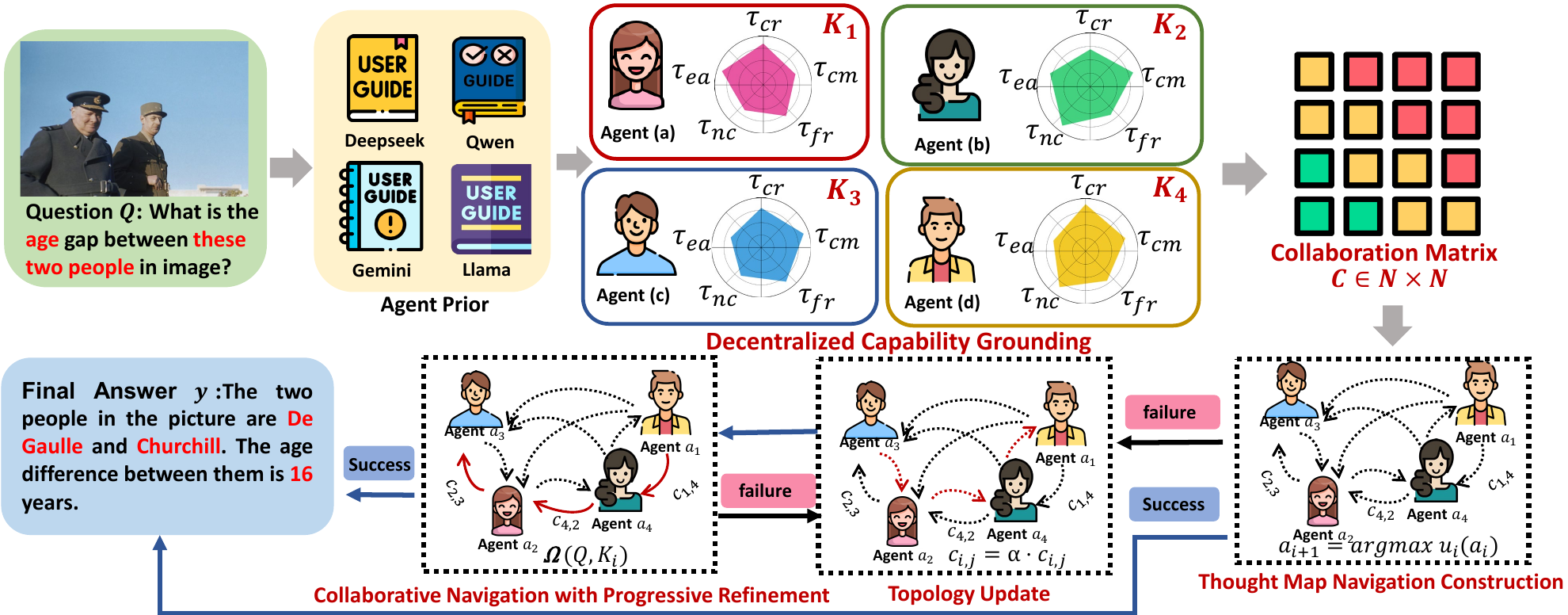}
    \caption{An overview of the \textbf{De}centralized \textbf{A}gentic \textbf{R}easoning framework. }
    \label{fig:2}
\end{figure*}
In this section, we present \textbf{De}centralized \textbf{A}gentic \textbf{R}easoning framework  in which agents autonomously decide when and with whom to collaborate during question answering. The execution of DeAR proceeds in three phases, corresponding to the three challenges identified in the introduction.

\subsection{Overview}

The core of DeAR is to navigate an \textit{Agentic Thought Map}, where nodes represent reasoning states and edges represent collaboration probabilities. This navigation is governed by a dynamic \textbf{Collaboration Propensity Matrix} $\mathcal{C} \in \mathbb{R}^{N \times N}$, which encodes the likelihood of beneficial interaction between agent pairs based on their grounded capabilities. As agents traverse the map, they generate a sequence of knowledge states $\mathcal{K}$. If a path leads to a dead end (i.e., fails to generate a valid answer $y$), the system employs a \textit{progressive refinement} strategy, updating the matrix $\mathcal{C}$ to prune invalid edges and steer the navigation toward effective reasoning chains. The system $\mathcal{S}$ is defined as:
\begin{equation}
\mathcal{S} = (Q, \mathcal{A}, \mathcal{C}, \mathcal{K}).
\end{equation}

As illustrated in Figure~\ref{fig:2}, the algorithm consists of three parts.

\subsection{Decentralized Capability Grounding}

To address the limitations of static role pre-definition, DeAR implements \textit{Decentralized Capability Grounding}. At initialization, agents are not assigned rigid labels (e.g., ``Math Agent''). Instead, each agent $a_i$ grounds its identity in verifiable linguistic benchmarks derived from its underlying official technical reports or model cards, denoted as $\mathcal{P}_i$.

Rather than relying on externally assigned task labels or groundless self-assessments, we characterize each agent’s capability along a set of fundamental cognitive dimensions commonly required in question answering. Specifically, each agent maintains an intrinsic capability profile:
\begin{equation}
\boldsymbol{\tau}_i =
\left[
\tau_i^{\text{cr}},
\tau_i^{\text{ea}},
\tau_i^{\text{nc}},
\tau_i^{\text{fr}},
\tau_i^{\text{cm}}
\right],
\end{equation}
where $\tau_i^{\text{cr}}$, $\tau_i^{\text{ea}}$, $\tau_i^{\text{nc}}$, $\tau_i^{\text{cm}}$,and $\tau_i^{\text{fr}}$ denote the agent’s proficiency in commonsense reasoning, explanatory ability, numerical computation, natural language comprehension ,and factual reliability, respectively. The detailed prompt templates and alignment protocols employed for this calibration mechanism are provided in Appendix.

Crucially, these continuous values do not originate from arbitrary text reading. They are based on explicit mappings of quantitative benchmark data extracted from $\mathcal{P}_i$. Let $s_i^{d}$ represent the raw accuracy score of agent $a_i$ on a standard dataset corresponding to dimension $d$. For instance, $s_i^{\text{nc}}$ is derived from the GSM8K benchmark, $s_i^{\text{cr}}$ from MMLU, and $s_i^{\text{fr}}$ from TruthfulQA. To resolve the inconsistency of reporting standards across different foundational models, we apply a calibration mechanism during the system initialization phase. The raw scores are projected into the $[0,1]$ interval and normalized across all $N$ agents using a Softmax operation:
\begin{equation}
\tau_i^{d} = \frac{\exp(s_i^{d})}{\sum_{k=1}^N \exp(s_k^{d})}, \quad d \in \{\text{cr}, \text{ea}, \text{nc}, \text{fr},\text{cm}\}.
\end{equation}
This mapping mechanism ensures that the capability values $\boldsymbol{\tau}_i$ of all agents are rigorously quantified and strictly comparable on the same scale.

Upon receiving a query $q$, agent $a_i$ dynamically contextualizes its profile based on the query requirements and its technical report $\mathcal{P}_i$, producing a \textbf{Query-Dependent Capability State}:
\begin{equation}
\boldsymbol{\tau}_i(q) = \Phi_i\left(q, \mathcal{P}_i, \boldsymbol{\tau}_i\right),
\end{equation}
where $\Phi_i(\cdot)$ denotes the agent’s internal reasoning operator, implemented via prompting, which evaluates the task context against the agent's calibrated benchmark profile.

Finally, the knowledge or intermediate reasoning generated by agent $a_i$ is conditioned on the query and this capability state:
\begin{equation}
\mathcal{K}_i =\Theta_i \left(q \mid \boldsymbol{\tau}_i(q)\right).
\end{equation}
Here, $\Theta_i(\cdot)$ denotes the LLM's generative reasoning process driven by its parametric knowledge, yielding the intermediate results $\mathcal{K}_i$. By evaluating peer benchmark reports, agent $a_i$ quantitatively estimates their strengths, forming the basis to compute the collaboration propensity $c_{i,j}$ and construct the decentralized thought map.
\subsection{Self-Organized Collaboration}
\label{sec:collab_path}

\begin{figure*}[tbp] 
    \centering
    \includegraphics[width=0.9\textwidth]{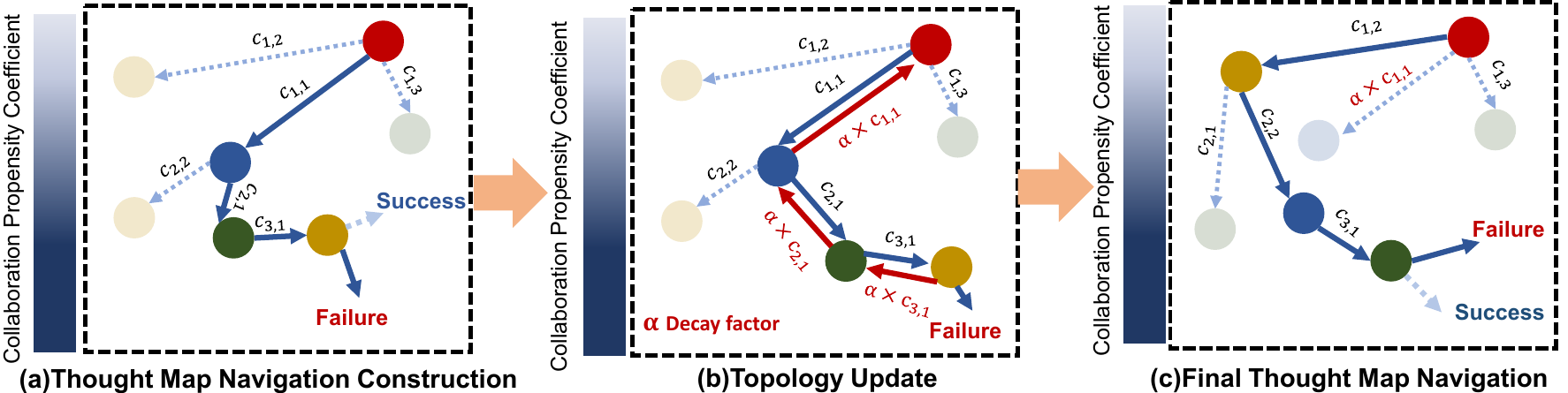}
    \caption{Thought Map Navigation Construction among Agents}
    \label{fig:3}
\end{figure*}

To replace centralized routing with local consensus, we introduce a mechanism for \textit{Self-Organized Collaboration}. Each agent $a_i$ autonomously evaluates the potential utility of peering with every other agent $a_j$ under the current query context.To initiate the decentralized reasoning trajectory, the first active agent is selected uniformly at random from the agent pool. This evaluation forms a Collaboration Propensity Vector $c_i$:
\begin{equation}
c_i =
\left[
c_{i,1},\,
c_{i,2},\,
\ldots,\,
c_{i,N}
\right],
\end{equation}
where $c_{i,j} \in [0,1]$ is the collaboration preference towards agent $a_j$. Given a query mapping to a cognitive dimension $d$, $c_{i,j}$ is calculated via a temperature-scaled Softmax over the calibrated proficiency $\tau_j^d$ defined in Equation 3:
\begin{equation}
c_{i,j} = \frac{\exp(\tau_j^d / T)}{\sum_{k=1, k \neq i}^N \exp(\tau_k^d / T)}, \quad \forall j \neq i
\end{equation}
where $T > 0$ controls routing sharpness. Since $\tau_j^d \in [0,1]$ is bounded, this formulation inherently prevents exponential overflow, guaranteeing absolute numerical stability. By definition, self-collaboration is disallowed:
\begin{equation}
c_{i,i} = 0, \quad \forall i \in \{1,2,\ldots,N\}.
\end{equation}

Aggregating the collaboration preference vectors from all agents yields a collaboration matrix $\mathcal{C}$:
\begin{equation}
\mathcal{C} =
\left(
\begin{array}{cccc}
0 & c_{1,2} & \cdots & c_{1,N} \\
c_{2,1} & 0 & \cdots & c_{2,N} \\
\vdots & \vdots & \ddots & \vdots \\
c_{N,1} & c_{N,2} & \cdots & 0
\end{array}
\right)
\end{equation}
This matrix encodes pairwise and directed collaboration preferences among agents, reflecting both agent heterogeneity and task-specific requirements. Unlike static routing strategies or centralized coordination mechanisms, the collaboration structure here is dynamically shaped by each agent’s local assessment under the current query. Although the base ranking of $c_{i,j}$ reflects target capabilities, it avoids global ranking degeneration by activating agents on an as-needed basis. The cumulative context $\mathcal{K}_i$ from operator $\Theta_i$ imbues the payload with source-specificity, achieving context-dependent collaboration.
\paragraph{Thought Map Navigation construction}

Based on the collaboration matrix $\mathcal{C}$, agents construct a collaboration path in a sequential manner. The objective of this process is to determine an ordered sequence of agents such that the knowledge generated by each selected agent is incrementally propagated and refined through interaction.

The navigation begins with a randomly selected seed agent (or the most relevant initial agent). At step $i$, the current agent $a_i$ acts as a local navigator, selecting the next node (agent $a_j$) that maximizes the collaboration propensity. The incremental utility of transitioning to an unvisited agent $a_j$ is:
\begin{equation}
u_i(a_j) =
\begin{cases}
c_{i,j}, & a_j \notin \mathcal{V}_i, \\
0, & a_j \in \mathcal{V}_i.
\end{cases}
\end{equation}
The next agent $a_{i+1}$ is chosen by
\begin{equation}
a_{i+1} = \arg\max_{a_j \in \mathcal{A}} u_i(a_j)
\end{equation}

The selected agent then generates its knowledge conditioned on the input query and the accumulated context formed by the union of knowledge generated by previously selected agents, i.e., $\bigcup_{k=1}^{i} \mathcal{K}_k$. This creates a chain of thought that exploits the strongest collaborative edges in the map.

Through sequential selection and knowledge propagation, agents autonomously construct a thought map that progressively integrates complementary reasoning capabilities. Independent of any centralized judge or predefined topology, the terminal agent generates the final answer $y$ directly from the fully accumulated knowledge $\mathcal{K}_N$.Dynamic Termination. This decentralized trajectory terminates dynamically based on real-time capability assessments rather than a fixed step limit. Navigation halts if the evaluated propensities of all unvisited peers (outside $\mathcal{V}_i$) fall below a minimal operational threshold, or if all agents have been exhausted. Upon halting, the active agent serves as the terminal node, outputting its accumulated context as the final answer.



\subsection{Collaborative Navigation with Progressive Refinement}



Upon reaching the terminal state, the system attempts to generate a final answer $y$. If the accumulated context $\mathcal{K}_N$ is insufficient (a ``dead end'' on the map), we do not simply discard the effort. Instead, we perform a \textbf{Topology Update} $\Psi(\cdot)$ to refine the thought map.
\begin{equation}
y = \Omega\left(q, \mathcal{K}_N\right),
\end{equation}
where $\Omega(\cdot)$ denotes the answer generation function of the terminal
agent $a_N$ conditioned on the full accumulated context.

Specifically,DeAR leverages the terminal agent rather than an external judge to verify the final context $\mathcal{K}_N$. If the final context  fails to match the initial query and its requirements, a refusal is triggered. To prune the target edge set $\mathcal{E}_{\text{fail}}$ without over-penalizing valid upstream transitions, DeAR employs a localized, progressive backtracking strategy instead of full-path mitigation. Initially, $\mathcal{E}_{\text{fail}}$ contains only the immediate terminal edge, prompting alternative routing at the current depth. If local choices are exhausted, $\mathcal{E}_{\text{fail}}$ expands step-by-step to preceding edges in the reasoning path. Finally, these active edges are penalized in the Collaboration Propensity Matrix:
\begin{equation}
\mathcal{C} \leftarrow \Psi(\mathcal{C}),
\end{equation}
where the transformation applies a decay factor $\alpha$ (e.g., 0.5) to the failed connections:
\begin{equation}
c_{i,j} =
\begin{cases}
\alpha \cdot c_{i,j}, & (a_i \rightarrow a_j) \in \mathcal{E}_{\text{fail}}, \\
c_{i,j}, & \text{otherwise}.
\end{cases}
\end{equation}
This update prunes failing edges, backtracking to preceding layers step-by-step if local routes fail. The agents then re-navigate the adjusted topology. This process of exploration, sequential backtracking, and refinement allows DeAR to progressively converge on a valid reasoning path, bypassing the single-point vulnerabilities typical of centralized systems.
\section{Experimental Setup}
\subsection{Datasets}
We evaluate our method on four widely used multimodal reasoning datasets: MMMU \citep{yue2023mmmu}, MathVista \citep{lu2024mathvista}, ChartQA \citep{hegde2025chartqa}, and ScienceQA \citep{lu2022learn}. We adopt Accuracy (ACC) as the primary evaluation metric and conduct experiments on the full test splits for all benchmarks. Specifically, beyond the overall accuracy for MMMU and ChartQA, we report detailed fine-grained metrics for MathVista across different question types (e.g., FQA, GPS, MWP) and reasoning skills (e.g., ALG, ARI, GEO). For ScienceQA, we report the accuracy across various subjects (NAT, SOC, LAN), modalities (TXT, IMG, NO), and grade levels (G1-6, G7-12), alongside the average score.

Furthermore, to demonstrate the generalizability of DeAR, we additionally conduct experiments on 5 text-based QA benchmarks, encompassing both single-hop (e.g., NQ\citep{kwiatkowski2019natural}, TriviaQA\citep{joshi2017triviaqa}, PopQA\citep{mallen2023not}) and multi-hop (e.g., WikiMultihopQA\citep{ho2020constructing}, and HotpotQA\citep{yang2018hotpotqa}) datasets. For these pure text tasks, we employ Exact Match (EM) and F1 as evaluation metrics, with 1,000 examples randomly sampled for the multi-hop evaluations.
\subsection{Implementation Details}
In our implementation, the thought map uses a dynamic multi-agent design where up to four agents can participate depending on the query's complexity. For multimodal reasoning tasks, we assign a different Multimodal Large Language Model (MLLM) to each agent: DeepSeek-VL-7B-Chat \citep{lu2024deepseekvl}, LLaVA-1.5-7B \citep{liu2023llava}, Qwen2-VL-7B-Instruct \citep{Qwen2VL}, and MiniCPM-V-2.6 \citep{yao2024minicpm}. Using different models provides diverse visual reasoning perspectives. 
For text-based QA evaluations, the agents are powered by four distinct LLMs (Qwen3-8B \citep{yang2025qwen3}, Gemma-3-1B-IT \citep{team2025gemma}, Llama-3.2-3B \citep{dubey2024llama}, and DeepSeek-LLM-7B-Chat \citep{liu2024deepseek}). We integrate these text models with the FlashRAG toolkit \citep{jin2025flashrag}, using E5-base-v2 as the dense retriever to fetch the top $k=10$ passages from a shared Wikipedia corpus. 
We keep all generation and system hyperparameters consistent across the multimodal and QA experiments. Finally, all models are evaluated in a zero-shot setting without any task-specific fine-tuning.

\begin{table*}[t]
\centering
\small
\renewcommand{\arraystretch}{1.2} 
\resizebox{1\linewidth}{!}{
\begin{tabular}{l | c | c c c c c c | c | c c c c}
\hline

\multirow{2}{*}{\textbf{Method}} & 
\multirow{2}{*}{\textbf{MMMU}} & 
\multicolumn{6}{c|}{\textbf{MathVista}} & 
\multirow{2}{*}{\textbf{ChartQA}} & 
\multicolumn{4}{c}{\textbf{ScienceQA}} \\
\cline{3-8} \cline{10-13}

& & FQA & GPS & MWP & TQA & VQA & \textbf{ALL} & & NAT & SOC & LAN & \textbf{ALL} \\
\hline 

\multicolumn{13}{c}{\textit{Baseline Models (Single Agent)}} \\
\hline 
LLaMA-2-13B \citep{liu2023llava}        & -     & 26.8 & 29.3 & 16.1 & 32.3 & 26.3 & 26.10 & -     & 44.1 & 41.2 & 43.9 & 43.08 \\
InternVL2.5-26B \citep{chen2024internvl}& 51.80 & 38.0 & 35.0 & 30.0 & 40.0 & 37.0 & 36.03 & -     & -    & -    & -    & - \\
MiniCPM-V-2.6 \citep{yao2024minicpm}    & 45.11 & 51.7 & 27.4 & 39.8 & 42.5 & 34.7 & 39.89 & -     & 59.5 & 57.0 & 59.6 & 58.70 \\
Qwen2-VL-72B \citep{Qwen2VL}            & 46.21 & 55.9 & 34.7 & 29.7 & 58.8 & 42.4 & 41.34 & 88.30 & -    & -    & -    & - \\
NVLM-H 1.0 78B \citep{nvlm2024}         & 53.08 & 65.0 & 48.0 & 45.0 & 68.0 & 53.0 & 55.91 & -     & -    & -    & -    & - \\
\hline 

\multicolumn{13}{c}{\textit{Baseline Models (Multi-Agent Frameworks)}} \\
\hline 
MAD-Vote (EMNLP 2024) \citep{liang2024encouraging} & 44.70 & 49.0 & 38.0 & 35.0 & 51.0 & 42.0 & 43.12 & 81.02 & 55.0 & 52.1 & 54.0 & 53.70 \\
MUG (AAAI 2025) \citep{liang2025multi}             & 50.30 & 55.0 & 40.0 & 39.0 & 56.0 & 46.0 & 47.32 & -     & 56.5 & 53.8 & 55.5 & 55.27 \\
C2R (EMNLP 2025) \citep{jang2025confidence}        & 51.92 & 60.1 & 38.2 & 45.4 & 62.0 & 49.3 & 51.01 & 85.08 & -    & -    & -    & - \\
Cache-of-Thought (EMNLP 2025) \citep{wu2025cache}  & 37.92 & 54.2 & 35.4 & 30.0 & 58.3 & 44.6 & 44.51 & -     & 59.0 & 56.5 & 59.8 & 58.44 \\
Corvid (ICCV 2025) \citep{jiang2025corvid}         & 52.20 & 65.0 & 40.8 & 35.7 & 70.6 & 50.0 & 52.43 & -     & -    & -    & -    & - \\
Insight-V (CVPR 2025) \citep{dong2025insight}      & 53.70 & 78.0 & 48.5 & 42.1 & 70.3 & 53.1 & 58.40 & 83.00 & 63.5 & 59.2 & 63.0 & 61.93 \\
AutoGen \citep{wu2024autogen}                       & 54.31 & 77.6 & 49.2  & 42.7 & 68.7 & 54.5 & 59.01 & 86.79  & 62.7   & 58.7  & 62.1 & 60.24 \\
AgentVerse \citep{chen2024agentverse}               & 55.10 & 76.4 & 47.1   & 41.6  & 69.1 & 54.4 & 58.79 & 87.12 & 61.9   & 58.4   & 62.9   & 61.45 \\
DyLAN \citep{liu2024dynamic}                           & 54.39 & 77.2 & 48.9  & 42.9  & 67.1  & 53.9 & 59.16 & 86.46 & 62.4    & 59.1   & 61.3  &60.52 \\
\hline 

\multicolumn{13}{c}{\textit{Our Proposed Method}} \\
\hline 
DeAR (Fixed Role)                       & 49.82 & 68.1 & 41.2 & 38.9 & 60.1 & 49.7 & 51.60 & 85.72 & 61.0 & 57.5 & 60.6 & 59.70 \\
DeAR (w/o Thought Map)                  & 47.20 & 68.2 & 41.3 & 39.0 & 60.2 & 49.7 & 51.68 & 83.12 & 55.5 & 51.2 & 55.5 & 54.07 \\
DeAR (w/o Topology Update)              & 53.21 & 75.0 & 45.0 & 41.0 & 70.5 & 46.7 & 55.65 & 88.01 & 61.2 & 57.5 & 61.0 & 59.92 \\

\textbf{DeAR (Ours)}                    & \textbf{55.58} & \textbf{80.3} & \textbf{49.9} & \textbf{42.7} & \textbf{84.5} & \textbf{60.9} & \textbf{59.41} & \textbf{89.43} & \textbf{64.8} & \textbf{60.3} & \textbf{62.2} & \textbf{62.45} \\

\hline
\end{tabular}
}
\caption{Zero-shot evaluation on four multimodal benchmarks (fine-grained metrics reported for MathVista and ScienceQA). '-' denotes unreported data. Agent counts: Cache-of-Thought (2); MUG, C2R (3); MAD-Vote, Corvid, Insight-V, AutoGen, AgentVerse, DyLAN, ReConcile, and DeAR (4). Best results in \textbf{bold}.}
\label{tab:master_results}
\end{table*}

\begin{table*}[!htbp]
\centering
\renewcommand{\arraystretch}{1.2}
\resizebox{\textwidth}{!}{%
\begin{tabular}{l|cccc|cccc|cccc|cccc|cccc}
\hline
\textbf{Method}
& \multicolumn{4}{c|}{\textbf{NQ}}
& \multicolumn{4}{c|}{\textbf{TriviaQA}}
& \multicolumn{4}{c|}{\textbf{PopQA}}
& \multicolumn{4}{c|}{\textbf{2Wiki}}
& \multicolumn{4}{c}{\textbf{HotpotQA}} \\
\cline{2-21}
 & EM & $\Delta$EM & F1 & $\Delta$F1
 & EM & $\Delta$EM & F1 & $\Delta$F1
 & EM & $\Delta$EM & F1 & $\Delta$F1
 & EM & $\Delta$EM & F1 & $\Delta$F1
 & EM & $\Delta$EM & F1 & $\Delta$F1 \\
\hline
\multicolumn{21}{c}{\textit{Single Agent Framework}} \\
\hline
Vanilla Gen
& 17.40 & +29.79 & 26.27 & +27.86
& 56.60 & +4.87 & 65.50 & +7.54
& 19.20 & +20.54 & 23.33 & +23.87
& 8.60 & +42.53 & 16.50 & +44.74
& 16.80 & +31.51 & 23.88 & +29.86 \\
\hline
\multicolumn{21}{c}{\textit{Multi-Agent Frameworks}} \\
\hline
MAD\citep{du2023improving}(ICML 2024)
& 21.80 & +25.39 & 33.11 & +21.02
& 56.40 & +5.07 & 66.39 & +6.65
& 21.40 & +18.34 & 28.28 & +18.92
& 18.20 & +32.93 & 25.13 & +36.11
& 23.00 & +25.31 & 32.79 & +20.95 \\
IRCoT\citep{trivedi2023interleaving}(ACL 2023)
& 28.60 & +18.59 & 37.36 & +16.77
& 47.20 & +14.27 & 54.56 & +18.48
& 27.00 & +12.74 & 33.02 & +14.18
& 32.80 & +18.33 & 31.19 & +30.05
& 25.20 & +23.11 & 34.40 & +19.34 \\
Iter-RetGen\citep{shao2023enhancing}(Arxiv 2023)
& 40.80 & +6.39 & \cellcolor{gray!20}52.31 & +1.82
& \cellcolor{gray!40}63.00 & -1.53 & \cellcolor{gray!20}72.23 & +0.81
& \cellcolor{gray!20}39.60 & +0.14 & 46.41 & +0.79
& 15.00 & +36.13 & 24.75 & +36.49
& 27.80 & +20.51 & 38.93 & +14.81 \\
FLARE\citep{jiang2023active}(EMNLP 2023)
& 19.40 & +27.79 & 27.68 & +26.45
& 53.60 & +7.87 & 63.05 & +9.99
& 21.60 & +18.14 & 24.35 & +22.85
& 9.20 & +41.93 & 20.13 & +41.11
& 16.60 & +31.71 & 23.74 & +30.00 \\
Self-RAG\citep{asai2024self}(Arxiv 2024)
& \cellcolor{gray!20}44.00 & +3.19 & 52.20 & +1.93
& 46.40 & +15.07 & 58.37 & +14.67
& 22.00 & +17.74 & 34.38 & +12.82
& 13.00 & +38.13 & 26.63 & +34.61
& 14.80 & +33.51 & 28.81 & +24.93 \\
DRAG\citep{hu2025removal}(ACL 2025)
& 36.80 & +10.39 & 50.38 & +3.75
& 60.80 & +0.67 & 69.93 & +3.11
& 38.60 & +1.14 & \cellcolor{gray!20}46.50 & +0.70
& 28.80 & +22.33 & 36.97 & +24.27
& 30.80 & +17.51 & 41.74 & +12.00 \\
DeAR (Ours)
& \cellcolor{gray!40}47.19 & -- & \cellcolor{gray!40}54.13 & --
& \cellcolor{gray!20}61.47 & -- & \cellcolor{gray!40}73.04 & --
& \cellcolor{gray!40}39.74 & -- & \cellcolor{gray!40}47.20 & --
& \cellcolor{gray!40}51.13 & -- & \cellcolor{gray!40}61.24 & --
& \cellcolor{gray!40}48.31 & -- & \cellcolor{gray!40}53.74 & -- \\
\hline
\end{tabular}%
}
\caption{Overall evaluation results of the proposed DeAR framework and other baselines on five text-only QA benchmarks.
$\Delta$EM and $\Delta$F1 denote the performance gap between DeAR and each baseline ($\Delta = \text{DeAR score} - \text{baseline score}$).
Darker gray cells mark the best performance, and lighter gray cells mark the second-best performance among all methods.}
\label{tab:qa_comparison_wide}
\end{table*}
\section{Experimental Results}
\subsection{Comparison with Baselines}
Evaluating DeAR against large single models shows that improvements come from the structured architecture instead of parameter scaling. Despite relying on a cooperative system of four base models, DeAR maintains a smaller total parameter count than 72B or 78B models while showing higher accuracy. Specifically, the score of DeAR is 18.07 points higher than Qwen2-VL-72B on MathVista (59.41 vs. 41.34), as shown in Table~\ref{tab:master_results}.  

Compared to recent chain of thought and multi agent frameworks such as MAD-Vote\citep{liang2024encouraging}, MUG\citep{liang2025multi}, C2R\citep{jang2025confidence}, Cache of Thought\citep{wu2025cache}, Corvid\citep{jiang2025corvid}, and Insight-V\citep{dong2025insight}, DeAR reports the highest overall scores on MathVista (59.41), ChartQA (89.43), and ScienceQA (62.45) in Table~\ref{tab:master_results}.It also consistently outperforms established multi agent platforms including AutoGen\citep{wu2024autogen}, AgentVerse\citep{chen2024agentverse}, and DyLAN\citep{liu2024dynamic} across all four benchmarks. For instance, DeAR surpasses these three baselines on ChartQA by over 2.3 and achieves a higher score on MMMU (55.58), showing the advantage of a decentralized topology over conventional centralized routing. 

Evaluations on five text QA benchmarks show the performance of DeAR beyond multimodal tasks, with detailed metrics provided in Table~\ref{tab:qa_comparison_wide}. The framework ranks first on NQ and obtains the highest F1 score on TriviaQA for single hop tasks. For multi hop queries, it obtains the highest score on HotpotQA and maintains stable precision and completeness on 2Wiki (Table~\ref{tab:qa_comparison_wide}), showing that decentralized thought navigation applies to pure NLP domains.  

Table~\ref{tab:master_results} decouples architectural gains from ensemble effects using identical backbones. DeAR (Fixed Role), serving as the centralized routing baseline, drops sharply (-5.76 on MMMU). DeAR (w/o Topology Update), acting as the ensemble control without backtracking, also degrades (-2.37 on MMMU). These margins mathematically prove our superiority stems from decentralized navigation rather than vanilla model aggregation.
\subsection{Analysis of the Thought Map Navigation}
This section evaluates the effectiveness of the proposed thought map navigation by comparing it against a variant without (w/o) the thought map navigation. To provide deeper insights beyond aggregate metrics, we conduct a fine-grained analysis of the performance across diverse question types and reasoning skills within the MathVista dataset. The results demonstrate that dynamic agent routing consistently outperforms static setups across all sub-categories, confirming that adaptive collaboration is essential for complex visual reasoning.

Overall, the thought map navigation increases the comprehensive MathVista accuracy from 51.68 to 59.41. The detailed breakdown reveals that the most substantial gains occur in tasks demanding intricate multi-step logic and specialized visual processing. For example, accuracy on Statistical Reasoning (STA) improves significantly from 62.9 to 80.9. Similarly, Logical Reasoning (LOG) sees a major boost from 19.6 to 31.0, and Figure Question Answering (FQA) rises from 69.1 to 80.3. These consistent enhancements validate that allowing agents to autonomously navigate the reasoning space is far more effective than forcing a predefined execution order.

\subsection{Analysis of Topology Update}

This section evaluates the proposed topology update on multimodal benchmarks, including MMMU, MathVista, ChartQA, and ScienceQA. By correcting intermediate errors, the topology update consistently improves accuracy across datasets with varying visual complexities. On MMMU, the score increases from 53.21 to 55.58, mitigating errors in college level academic tasks, while MathVista accuracy rises from 55.65 to 59.41 in intricate mathematical settings. Similar gains are observed on ChartQA (88.01 to 89.43) and ScienceQA (59.92 to 62.45), reflecting a reduced accumulation of multiple step reasoning errors.
\paragraph{\textbf{Effect of Decay Factor $\alpha$}}
\begin{figure}
    \centering  
    \includegraphics[width=0.9\linewidth]{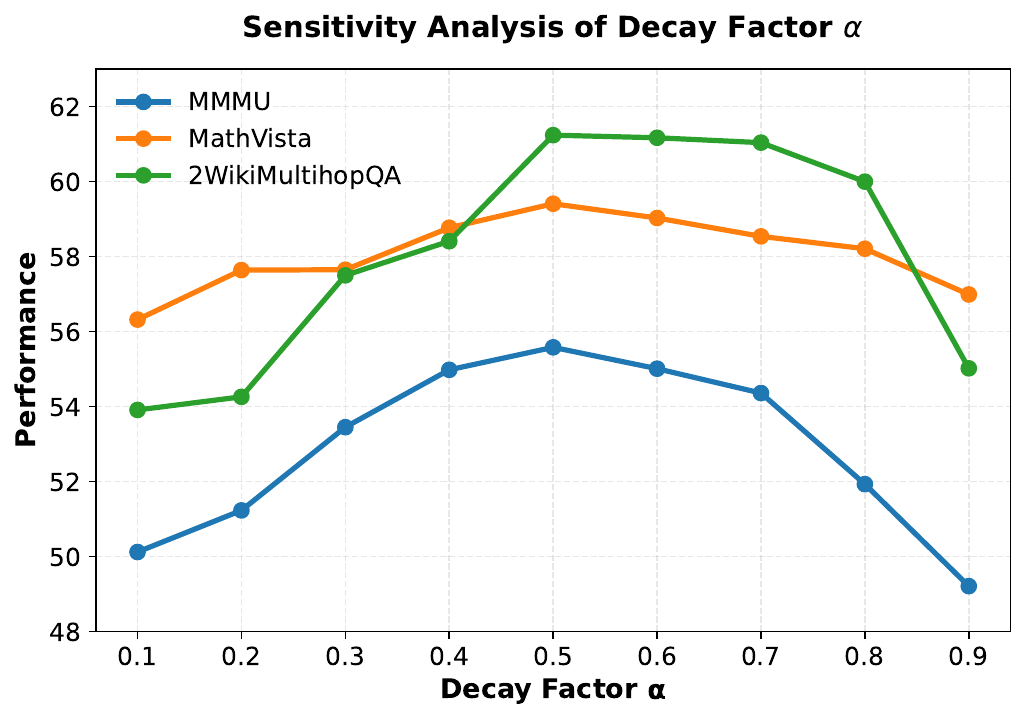}
    \caption{Analysis of the topology update's decay factor $\alpha$.} 
    \label{fig:8}     
\end{figure}
Figure \ref{fig:8} investigates the topology update's decay factor $\alpha \in [0.1, 0.9]$, which controls the penalty for failed paths. Small values ($\alpha=0.1$) cause overcorrection and unstable thought map paths, dropping the 2Wiki F1 score to 53.91. Conversely, large values ($\alpha=0.9$) insufficiently suppress ineffective paths, causing repeated invalid computations and a 2Wiki F1 of 55.02. DeAR performs optimally in the $[0.5,0.8]$. Specifically, a moderate $\alpha=0.5$ achieves peak performance on MMMU (55.58) and MathVista (59.41) by effectively selecting new paths without discarding previously successful interactions.

\begin{figure*}[tbp] 
    \centering
    \includegraphics[width=1\textwidth]{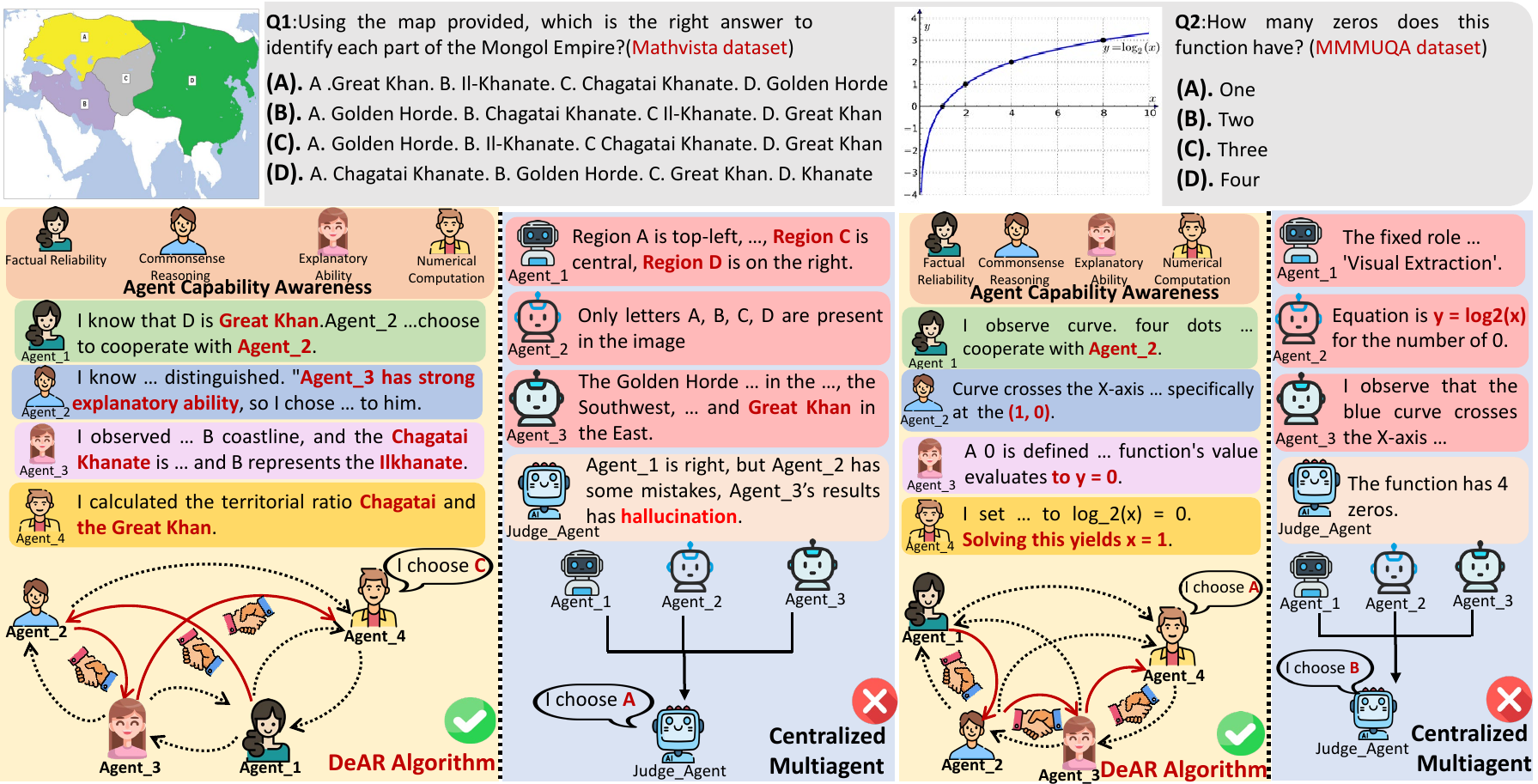}
    \caption{Case Study of the Reasoning with Centralized and Decentralized Multi-Agent Collaboration on Q\&A.}
    \label{fig:9}
\end{figure*}
\subsection{Dynamic Agent Activation and Usage Frequency}
DeAR constructs reasoning paths dynamically based on query requirements and agent confidence. The framework does not mandate the participation of all $N=4$ agents; unselected agents remain inactive if a smaller subset can resolve the query, which avoids unnecessary compute consumption. Table~\ref{tab:agent_frequency} presents the activation frequency of different agent subset sizes on MathVista. The results show that while complex problems require all 4 agents (37.5\%), a majority of queries (62.5\%) are resolved by fewer agents, confirming the adaptability and resource efficiency of the algorithm.
\begin{table}[ht]
\centering
\caption{Frequency of activation for varying numbers of agents on the MathVista dataset.}
\label{tab:agent_frequency}
\begin{tabular}{l c}
\toprule
\textbf{Number of Agents Activated} & \textbf{Frequency (\%)} \\
\midrule
1 Agent  & 14.3 \\
2 Agents & 21.8 \\
3 Agents & 26.4 \\
4 Agents & 37.5 \\
\bottomrule
\end{tabular}
\end{table}



\subsection{Case Study}
Figure \ref{fig:9} compares DeAR and a centralized multi-agent baseline on historical map identification (Q1) and chart reasoning (Q2).DeAR resolves visual and logical ambiguities through dynamic, capability-driven hand-offs. In Q1, agents progressively determine map assignments by combining factual knowledge, spatial commonsense, and historical geographic constraints, such as identifying landlocked regions. They finalize the boundaries via numerical verification of area ratios. In Q2, rather than being misled by the four plotted dots, the agents translate the mathematical "zero" into a geometric target on the x-axis and perform an algebraic check ($\log_2(x)=0 \Rightarrow x=1$) to confirm the single root.The centralized framework restricts agents to isolated, static roles like visual extraction or text reading. All intermediate findings are forced through a single Judge Agent, creating an information bottleneck. Without peer-to-peer verification, the Judge blindly aggregates fragmented reports. As a result, it hallucinates an incorrect map configuration in Q1 by forcibly merging vague hints. In Q2, the Judge directly misattributes the visual worker's observation of "four prominent dots" to the mathematical query and incorrectly outputs "Four".These results confirm that decentralized navigation effectively resolves complex logical dependencies and prevents the aggregation errors typical of centralized pipelines.
\section{Conclusion}
We propose \textbf{DeAR}, a decentralized framework replacing centralized multi-agent coordination with autonomous, capability-aware collaboration. DeAR integrates three core mechanisms: Decentralized Capability Grounding for query-dependent agent specialization, Thought Map Navigation for selective peer interaction, and Topology Update for adaptive error correction. These components allow agents to progressively refine reasoning trajectories without a central judge. Evaluations across 9 diverse multimodal and text-based QA benchmarks show DeAR consistently outperforms recent baselines, validating that decentralized, adaptive collaboration among agents enhances performance in complex tasks.

\bibliography{aaai2027}


\end{document}